\documentclass[runningheads]{llncs}
\usepackage{graphicx}

\usepackage{graphicx}
\usepackage{amsfonts}
\usepackage{amsmath}
\usepackage{bm}
\usepackage{mathtools}
\usepackage{hyperref}
\usepackage{caption}
\usepackage{subcaption}
\usepackage{booktabs}
\usepackage[misc]{ifsym}
\usepackage{makecell}
\usepackage{diagbox}
\usepackage{pifont}
\usepackage{colortbl}
\usepackage{multirow}
\usepackage{lipsum}

\usepackage{amssymb}
\usepackage{latexsym}

\usepackage{siunitx}

\usepackage{url}
\usepackage{xcolor}

\usepackage{hyperref}

\definecolor{newcolor}{rgb}{.8,.349,.1}
\definecolor{Gray}{gray}{0.85}
\definecolor{LightGray}{gray}{0.95}

\newcolumntype{a}{>{\columncolor{Gray}}c}
\newcolumntype{b}{>{\columncolor{LightGray}}c}

\newrobustcmd{\B}{\bfseries}
\newrobustcmd{\U}{\underline}
\newcommand{\xmark}{\ding{55}}
\newcommand{\cmark}{\ding{51}}
\newcommand{\greycell}{\cellcolor{lightgray}}

\usepackage{orcidlink}

\begin{document}
\title{Synthetic Privileged Information Enhances Medical Image Representation Learning}
\titlerunning{Synthetic Privileged Information Enhances Representation Learning}

\definecolor{orcidlogocol}{HTML}{A6CE39}
\newcommand{\lforcid}{\orcidlink{https://orcid.org/0009-0003-3667-2001}}
\newcommand{\cworcid}{\orcidlink{https://orcid.org/0000-0002-6890-2401}}
\newcommand{\riorcid}{\orcidlink{https://orcid.org/0000-0003-4898-040X}}
\newcommand{\kyorcid}{\orcidlink{https://orcid.org/0000-0002-2318-1460}}

\author{Lucas Farndale\inst{1,2~}\textsuperscript{\Letter}\lforcid \and Chris Walsh\inst{1,2}\cworcid \and 
Robert Insall\inst{1,2,3}\riorcid \and
Ke Yuan\inst{1,2}\kyorcid}
\authorrunning{L. Farndale et al.}
%
\institute{University of Glasgow, Glasgow, Scotland, UK \\
\and
Cancer Research UK Scotland Institute, Glasgow, Scotland, UK \\
\and
University College London, London, England, UK \\
\email{lucas.farndale@glasgow.ac.uk}}
\maketitle              
\begin{abstract}
Multimodal self-supervised representation learning has consistently proven to be a highly effective method in medical image analysis, offering strong task performance and producing biologically informed insights. However, these methods heavily rely on large, paired datasets, which is prohibitive for their use in scenarios where paired data does not exist, or there is only a small amount available. In contrast, image generation methods can work well on very small datasets, and can find mappings between unpaired datasets, meaning an effectively unlimited amount of paired synthetic data can be generated. In this work, we demonstrate that representation learning can be significantly improved by synthetically generating paired information, both compared to training on either single-modality (up to $4.4\times$ error reduction) or authentic multi-modal paired datasets (up to $5.6\times$ error reduction).

\end{abstract}

\section{Introduction}

Recent work \cite{farndale2023trident,girdhar2023imagebind,radford2021learning} has established the efficacy of integrating multiple sources of data to improve representation learning for histopathology. It has been demonstrated that integrating additional information, either advanced, high content information or something more routine, can greatly enhance the features that are learned by models, as it can act as an additional source of implicit supervision. This is particularly notable for privileged information -- information that is available during training but not inference -- which can be distilled into models to greatly improve single-modality downstream task performance where the privileged information is not available. Paired datasets can be used to distil knowledge from the privileged data to the primary data, producing much more biologically informative representations and improving performance on a variety of a priori unseen tasks during evaluation by only training a small classifier head \cite{farndale2023trident}.

\begin{figure}[t]
    \centering
    \begin{subfigure}{0.5\textwidth}
        \includegraphics[width=\textwidth]{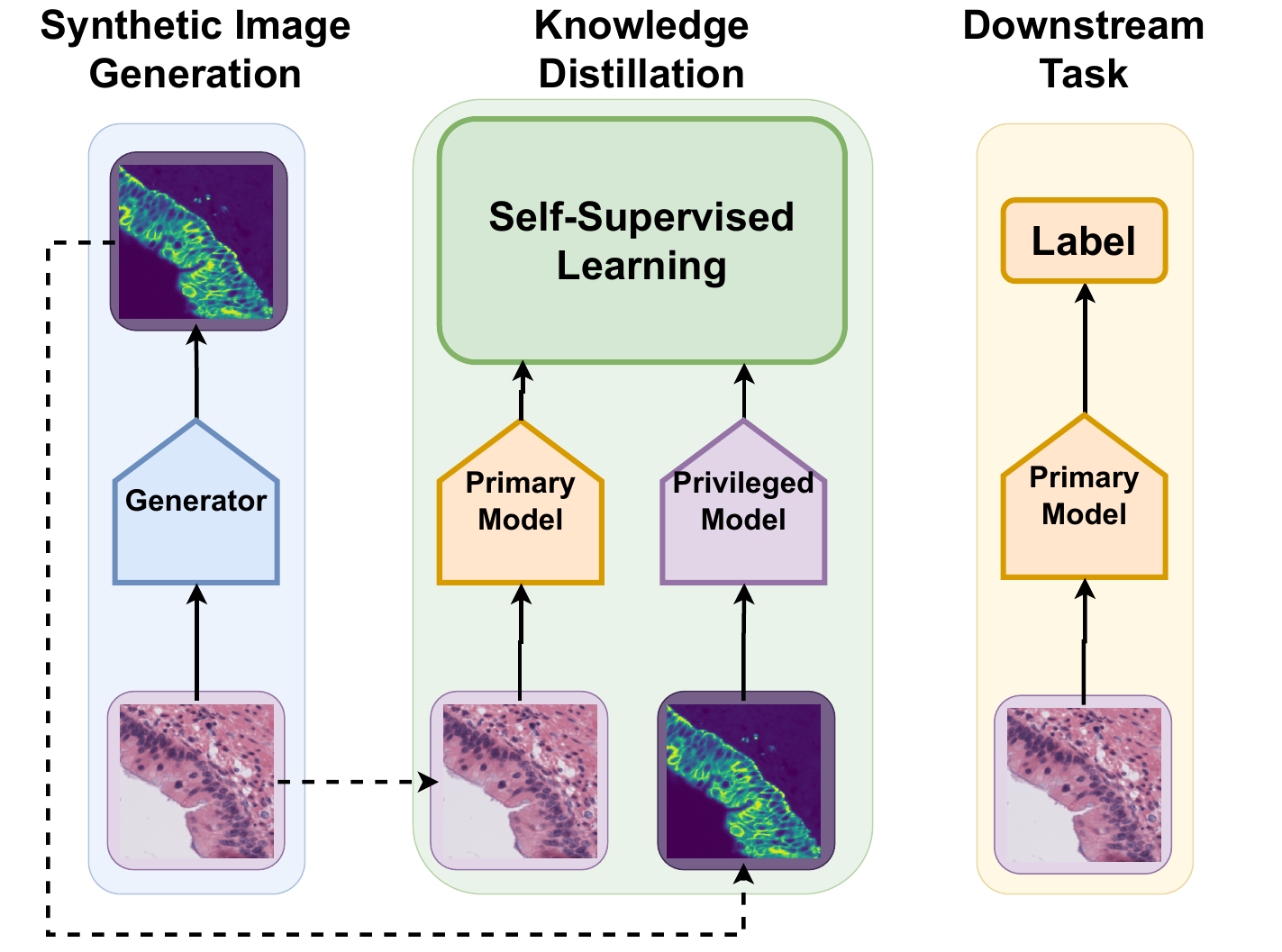}
        \caption{}
        \label{fig:pipeline}
    \end{subfigure}
    \begin{subfigure}{0.34\textwidth}
        \includegraphics[width=\textwidth, trim={ -3cm 0 0 0 }, clip]{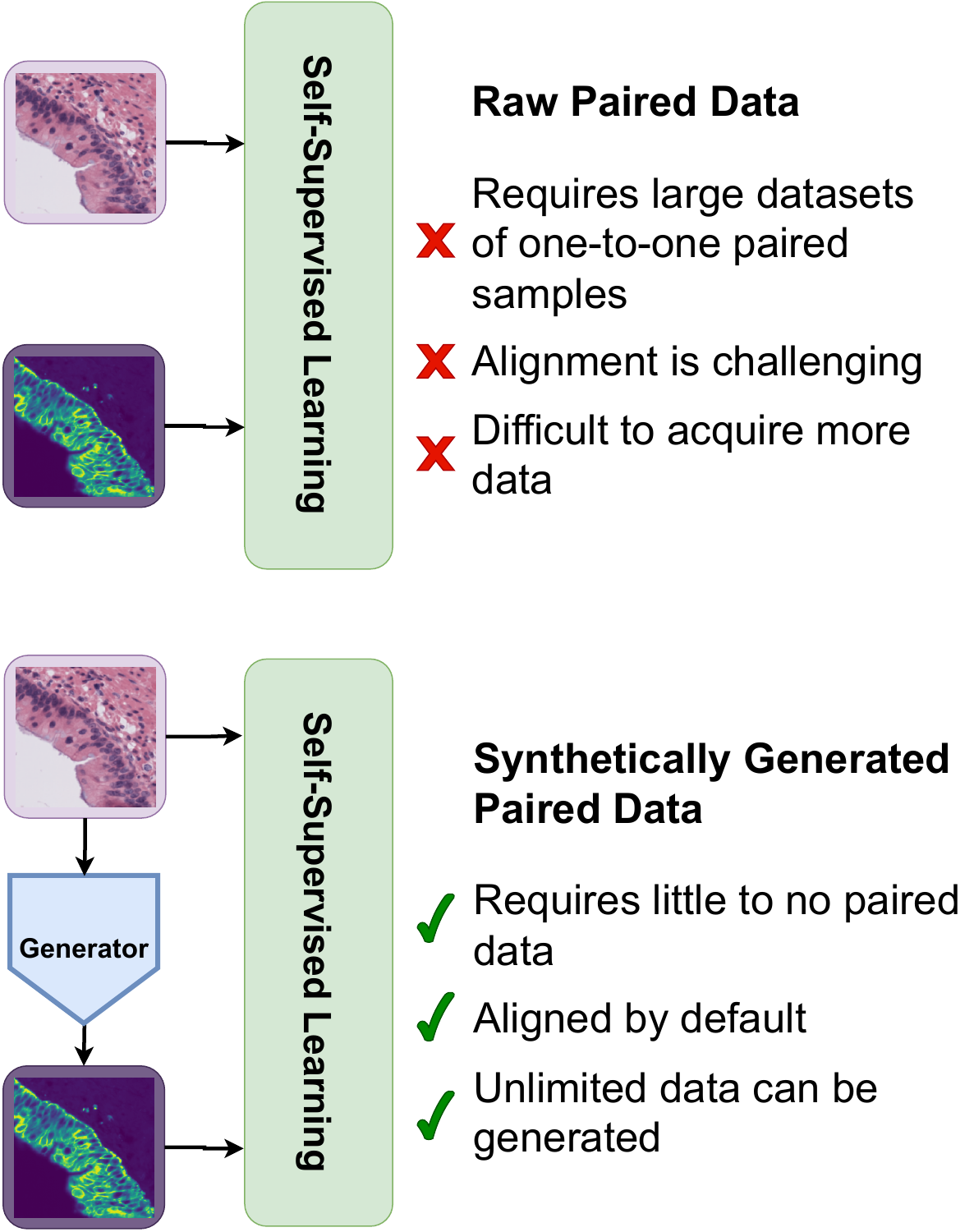}
        \caption{}
        \label{fig:examples}
    \end{subfigure}
    \caption{(a) Schematic of synthetic paired data being generated and passed to the self-supervised model (b) Comparison of standard multimodal self-supervised approaches against our proposed method using synthetically generated data. Images from \cite{burlingame2020shift}.}
    \label{}
\end{figure}

A major drawback of these approaches, however, is the reliance on large datasets where data are paired one-to-one during training. In many cases, such as histology, the privileged data of interest only exist in small datasets or in distinct, unpaired datasets which have not been previously amenable to these approaches. Image generation methods typically require orders of magnitude less data than self-supervised methods, with high quality image generation being possible with hundreds of examples (e.g. \cite{burlingame2020shift}), compared to hundreds of thousands for self-supervised learning \cite{el2021large}. Furthermore, cycleGAN \cite{zhu2017unpaired} does not require a one-to-one pairing of images and can achieve excellent results for image generation. We may therefore construct arbitrarily large, paired datasets using models which were trained on few or no paired samples.

In this work, we demonstrate that the benefits of training on large, paired datasets can be replicated using synthetically generated paired data for privileged self-supervised training. We show that high quality, biologically informed self-supervised models can be trained that benefit from privileged information, but only require a single data source for training and inference by using that input to generate a synthetic privileged pair. The key contributions of this work are:
\begin{itemize}
    \item We demonstrate that synthetically generated data can be used as privileged information in place of authentic source data, including where there is little or no real paired data, to improve model performance;
    \item We show that these models are more robust to distribution shift than unprivileged models, and encode meaningful biological features;
    \item We demonstrate that models trained with synthetic data closely mimic the representations learned from real data, and learn extra information compared to training with a single modality;
    \item We demonstrate the extra benefit from distilling synthetically generated data when the real dataset size is limited.
\end{itemize}

\section{Background and Related Work}

\textbf{Self-Supervised Learning} (SSL) is a midpoint between supervised and unsupervised, where inherent structural properties of data are leveraged to to create inductional biases which enable the model to learn without labels. The goal of this is to produce a vector representation of the image which has a much lower dimension than the original image but contains the useful semantics of the image needed for downstream applications. Recent work has extensively applied SSL in a wide variety of domains and tasks \cite{jing2020self}, and state-of-the-art SSL models have been shown to outperform both supervised and unsupervised models in many settings.

\textbf{Privileged Information} is any information that is available during training but not during inference. In a medical imaging context, this is true of an enormous corpus of scientific data, which has already been generated in one condition, but which would be impractical or expensive to generate again for future studies or to create a larger dataset. We are primarily motivated by histopathology, where vast quantities of brightfield H\&E stain datasets are available, but comparatively few for immunohistochemistry (IHC) stains and fewer again for state-of-the-art data such as spatial -omics, multiplex IHC, super-resolution microscopy or expert pixel-level annotations.

As Siamese networks will generally learn to encode only features which are shared between branches, if the inputs are different then only semantic content shared between inputs is learned. This means that useful features can be neglected if they are not shared. For example, it was shown that pairing thresholded immunofluorescence (IF) images with H\&E images significantly decreased model downstream performance on the H\&E images because the model neglected many details, such as fine-grained tissue structure, in these images that were not detectable in the IF, which were mostly black \cite{farndale2023trident}.

\textbf{TriDeNT} \cite{farndale2023trident} was developed to address this issue by  only neglecting features that are not present in the primary image. This means that features may still be neglected by the encoder of the privileged input, but only the encoder of the primary input is of interest for downstream tasks. TriDeNT works by utilising an additional primary input branch to generate a third representation, meaning that there are three concurrent joint-embedding latent spaces. The model must then balance retaining features that are present in both primary branches with those shared by the privileged information, and was shown to retain both feature sets, significantly improving downstream performance.

\textbf{Image Generation} methods have been extensively applied in medical imaging, both generating realistic images from a random seed \cite{kazerouni2023diffusion}, or virtually translating between modalities \cite{lyu2022conversion} or stains \cite{walsh2022ensuring,de2021deep}. Some image methods require paired data, such as pix2pix \cite{isola2017image}, but others, such as cycleGAN \cite{zhu2017unpaired}, can find mappings between unpaired datasets. This is especially useful in the context of histology, as the vast majority of additional stains are not from the same slice as their paired H\&E stains, and often are from different sections of the tissue block. This means it is hard to determine which parts of the images should be aligned when comparing different stains, as their morphologies can be very different. It has been shown that CycleGANs can be used to generate realistic IHC stains directly from H\&E stains without requiring corresponding input/target pairs \cite{xu2019gan,walsh2022ensuring}, and we demonstrate here that these synthetically generated stains can be used to improve the performance of self-supervised models using TriDeNT. These stains usually have the advantage of being perfectly spatially registered and aligned.

It has been shown that the representations from generative models perform poorly when evaluated on downstream tasks \cite{farndale2023more}, with even support vector machines outperforming state-of-the-art generative models \cite{claudio2021adversarial}. This is due to the models being highly optimised for the specific task of image generation, and therefore having representations which contain information, such as precise spatial locations of features, which are not useful for downstream tasks such as tissue type classification. It is therefore essential for representation learning to distil the useful information from these models' representations without over-constraining the model to replicate all of the features present, which reduces the quality of representations.

\begin{figure}[!h]
    \centering
    \begin{subfigure}{0.5\textwidth}
        \includegraphics[width=\textwidth]{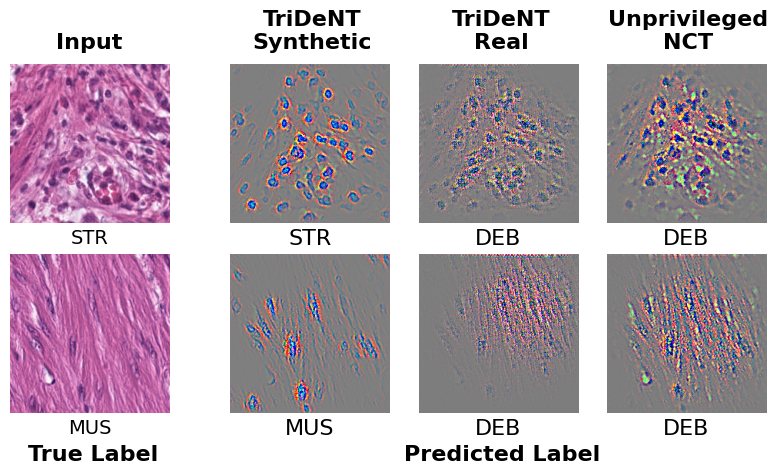}
        \caption{}
        \label{fig:saliency-maps}
    \end{subfigure}
    \begin{subfigure}{0.4\textwidth}
        \includegraphics[width=\textwidth, trim={ 0 -2.5cm 0 0 }, clip]{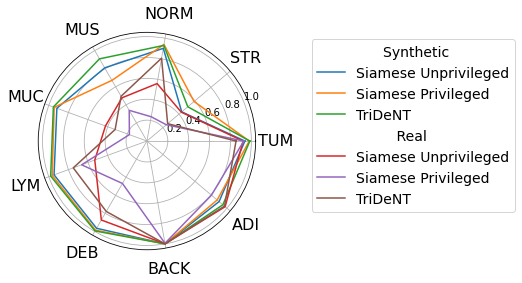}
        \caption{}
        \label{fig:radial_plot}
    \end{subfigure}

    \begin{subfigure}{0.9\textwidth}
        \scalebox{0.7}{\begin{tabular}{ccccccccccccc}
    \toprule
    & & & \multicolumn{2}{c}{PanNuke} & \multicolumn{2}{c}{IHC (cycleGAN)} & \multicolumn{2}{c}{IHC (pix2pix)} & \multicolumn{2}{c}{SHIFT} \\
    Loss & Method & Privileged & NCT & Camelyon & NCT & Camelyon & NCT & Camelyon & NCT & Camelyon \\
    \midrule
    \multirow{4}{*}{VICReg} & TriDeNT & \greycell\cmark & \greycell\B 0.9359~ & \greycell \B 0.9149 & \greycell\B 0.9252 & \greycell \U{0.9263} & \greycell \U{0.9184} & \greycell \U{0.8960} & \greycell\B 0.9383 & \greycell 0.6294 \\
    & \multirow{3}{*}{Siamese} & \greycell\cmark & \greycell\U{0.9044}~ & \greycell\U{0.8952} & \greycell \U{0.9234} & \greycell\B 0.9462 & \greycell\B 0.9250 & \greycell\B 0.8994 & \greycell \U{0.9236} & \greycell 0.5975 \\
    & & \xmark ~NCT & 0.8937~ & 0.7056 & 0.8937 & 0.7056 & 0.8937 & 0.7056 & 0.8937 & \B 0.7056 \\
    & & \xmark ~Real & $0.7301^\dag$ & 0.4636 & 0.8434 & 0.5360 & 0.8434 & 0.5360 & 0.8008 & \U{0.6387} \\
    \midrule
    \multirow{4}{*}{InfoNCE} & TriDeNT & \greycell\cmark & \greycell\B 0.9287~ & \greycell \U{0.8719} & \greycell\B 0.9331 & \greycell \U{0.9033} & \greycell\B 0.9302 & \greycell \U{0.8992} & \greycell \U{0.9115} & \greycell \U{0.6430} \\
    & \multirow{3}{*}{Siamese} & \greycell\cmark & \greycell\U{0.9103}~ & \greycell\B 0.8889 & \greycell \U{0.9260} & \greycell\B 0.9356 & \greycell \U{0.9259} & \greycell\B 0.9224 & \greycell\B 0.9299& \greycell 0.6007 \\
    & & \xmark ~NCT & 0.9047~ & 0.6205 & 0.9047 & 0.6205 & 0.9047 & 0.6205 & 0.9047 & 0.6205 \\
    & & \xmark ~Real & $0.7199^\dag$ & 0.5852 & 0.8623 & 0.5600 & 0.8623 & 0.5600 & 0.7921 & \B 0.7056 \\
    \midrule
    CrossEntropy & Supervised & \xmark & 0.9245 & 0.8440 & 0.9245 & 0.8440 & 0.9245 & 0.8440 & 0.9245 & 0.8440 \\
    \bottomrule
    \end{tabular}}
    \caption{}
    \label{fig:results}
    \end{subfigure}

    \begin{subfigure}{0.99\textwidth}
        \scalebox{0.63}{\begin{tabular}{ccccccccccccccccccccc}
    \toprule
    & & & \multicolumn{4}{c}{PanNuke} & \multicolumn{4}{c}{IHC (cycleGAN)} & \multicolumn{4}{c}{SHIFT} \\
    & & & \multicolumn{2}{c}{NCT} & \multicolumn{2}{c}{Camelyon} & \multicolumn{2}{c}{NCT} & \multicolumn{2}{c}{Camelyon} & \multicolumn{2}{c}{NCT} & \multicolumn{2}{c}{Camelyon} \\
    Loss & Method & Privileged & Synthetic & Real & Synthetic & Real & Synthetic & Real & Synthetic & Real & Synthetic & Real & Synthetic & Real \\
    \midrule
    \multirow{2}{*}{VICReg} & TriDeNT & \cmark & \greycell\B 0.9359 & $0.7337^\dag$ & \greycell\B 0.9149 & 0.5306 & \greycell\B 0.9252 & 0.8650 & \greycell\B 0.9263 & 0.6266 & \greycell\B 0.9383 & 0.7624 & \greycell\B 0.6294 & 0.5339 \\
    & Siamese & \cmark & \greycell\B 0.9044 & $0.6000^\dag$ & \greycell\B 0.8952 & 0.4831 & \greycell\B 0.9234 & 0.8256 & \greycell\B 0.9462 & 0.8614 & \greycell\B 0.9236 & 0.7439 & \greycell\B 0.5975 & 0.5120 \\
    \midrule
    \multirow{2}{*}{InfoNCE} & TriDeNT & \cmark & \greycell\B 0.9287 & $0.7530^\dag$ & \greycell\B 0.8719 & 0.6908 & \greycell\B 0.9331 & 0.8722 & \greycell\B 0.9033 & 0.8304 & \greycell\B 0.9115 & 0.7998 & \greycell\B 0.6430 & 0.5653 \\
    & Siamese & \cmark & \greycell\B 0.9103 & $0.5289^\dag$ & \greycell\B 0.8889 & 0.7134 & \greycell\B 0.9260 & 0.8370 & \greycell\B 0.9356 & 0.8423 & \greycell\B 0.9299 & 0.7464 & \greycell 0.6007 & \B 0.6601 \\
    \bottomrule
    \end{tabular}}
    \caption{}
    \label{fig:realvsynthetic}
    \end{subfigure}
    
    \caption{(a) Guided-GradCAM \cite{selvaraju2017grad} maps for samples from the NCT dataset. Predicted labels for each classifier are shown, as is the true label. Labels are adipose (ADI), background (BACK), debris (DEB), lymphocytes (LYM), mucus (MUC), muscle (MUS), normal mucosa (NORM), stroma (STR), and tumour (TUM). See Figure \ref{fig:full-guided-gradcam} for examples of all classes; (b) Breakdown of performance on the NCT evaluation task by class; (c) Results for models trained on the NCT dataset with synthetically generated privileged information as pairs for the NCT patches. Values marked with ``\dag'' are from \cite{farndale2023trident}; (d) Classification performance on NCT and Camelyon of models trained on the PanNuke/IHC/SHIFT datasets compared to the synthetically generated images paired with the NCT dataset.}
    \label{fig:results_fig}
\end{figure}

\section{Methodology and Datasets}

In all tasks we follow the implementation details of \cite{farndale2023trident}. We compare TriDeNT to Siamese baselines, with and without privileged information, and also to supervised learning. Encoders are ResNet-50 \cite{he2016deep} trained for 100 epochs with a warmup-cosine learning rate, with maximum value $10^{-4}$. Three-layer dense projection heads of width 8192 with batch normalisation were used, and classifier heads were single dense layers with softmax activations. For full details of datasets used, see Table \ref{tab:datasets}. We employ three evaluation tasks: tissue classification on the NCT dataset \cite{kather_dataset}, metastasis detection on the Camelyon17 dataset \cite{bandi2018detection}, and malignancy detection on the PanNuke dataset \cite{gamper2019pannuke}. We use existing image generation models trained on PanNuke, SHIFT \cite{burlingame2020shift} and a pix2pix model and a cycleGAN both trained on the IHC dataset detailed in \cite{walsh2022ensuring}. The pix2pix model uses paired image-to-image translation, while the cycleGAN uses unpaired images, allowing the necessity for paired data in model training to be assessed.

\section{Results}

\subsection{Synthetically Generated Segmentations as Privileged Data Improves Downstream Classification Performance}

In \cite{farndale2023more}, it was shown that standard Siamese unprivileged models failed to learn sub-nuclear features from histopathology images. This was addressed by pairing these images with machine-generated nuclear segmentations, which prompted the model to learn features related specifically to the nuclei. This was then further explored in \cite{farndale2023trident}, where it was shown that training models using TriDeNT on the Pannuke dataset \cite{gamper2019pannuke} with the segmentations as privileged information greatly improves their downstream task performance. Here we use the setting of \cite{farndale2023more}, generating segmentation maps using HoVer-Net \cite{graham2019hover}, and study the effect of using this as privileged information for TriDeNT. Note that PanNuke contains train/validation/test folds of approximately 2000 images each, meaning the SSL training dataset size is substantially increased by generating synthetically generated data from the NCT dataset.

The results in Figure \ref{fig:results_fig} demonstrate that distilling information from the synthetically generated nuclear segmentations greatly improves the performance of the model, and synthetically generated privileged information is found to be effectively distilled into the primary models. Notably, unprivileged learning on PanNuke or the NCT dataset performed poorly at distinguishing stroma from muscle, which requires subtle biological features to be learned. Models with synthetically generated privileged information performed considerably better. Encoders trained with privileged information retain features of both the background/connective tissue (using information from only the H\&E patches) and the nuclei (using information from both the H\&E patches and the segmentation masks), enabling them to make accurate predictions across different tissue types. Figures \ref{fig:saliency-maps} and \ref{fig:full-guided-gradcam} demonstrate that the TriDeNT model with synthetically generated privileged information learns to accurately identify the morphology of the nuclei in the images, making it easier to identify the phenotypic change between muscle and stroma. It was previously observed that unprivileged Siamese models tend to focus on the background, rather than the potentially more informative features of the nuclei \cite{farndale2023trident}. Privileged training with segmentation masks produced models which focused more on nuclei, this is equally true of synthetically generated privileged information. We find that even providing only the segmented nuclei as privileged information achieves improvements, although to a lesser extent than also providing the type annotations, as shown in Table \ref{tab:3_branch_image_nuclei}.

We note that there is a marked performance improvement for models with synthetically generated privileged information on the Camelyon17 task, which assesses the ability of the model to both detect metastasis and transfer to out-of-distribution datasets. Performance on this task is a good indicator of whether meaningful, generalisable biological features have been learned, as opposed to features specific to the train set or simply the contours of an image. It is known that privileged information causes in models to learn representations which are less prone to overfitting on small changes in irrelevant features such as colour, brightness, or blur, which can easily arise between datasets \cite{farndale2023trident}. Instead, models must focus on more useful semantic features shared between primary and privileged branches, such as the nuclei in Figures \ref{fig:saliency-maps} and \ref{fig:full-guided-gradcam}, which naturally gives rise to representations which transfer well to out-of-distribution data. We find that this is equally true for synthetic privileged information.

We also find that training on synthetically generated training sets offers significant improvements compared to using privileged information from the real source dataset, as shown in Figure \ref{fig:realvsynthetic}. This is due to the synthetic dataset being either larger (compared to PanNuke and SHIFT) or being more curated to encompass a distribution of samples (compared to IHC). This demonstrates that not only is using synthetic privileged information a good method to enhance performance on tasks related to the dataset the synthetic privileged information is generated \emph{for}, but it can also offer better results on tasks relating to the dataset the synthetic privileged information was generated \emph{from}.

As above, we see significant performance improvements for both pix2pix and cycleGAN models, showing that no real paired data is needed to achieve good results. It is notable that the synthetic SHIFT dataset improves performance on the NCT task, although this does not generalise well to the Camelyon evaluation task. The SSL performance is improved in spite of the generator making many verifiable errors, such as predicting fluorescence in patches labelled ``background'' or ``debris'', which makes little biological sense. This demonstrates that the addition of any useful privileged information can enhance performance, and synthetically generated data does not need to be perfect, with even models trained on very small datasets being suitable.

\begin{figure}[t]
    \centering
    \begin{subfigure}{0.5\textwidth}
        \scalebox{0.85}{
    \begin{tabular}{cccccc}
        \toprule
        Supervised & Method & Privileged & Synthetic & Real \\
        \midrule
        \multirow{3}{*}{\xmark} & TriDeNT & \cmark & \B 0.7508 & 0.5545 \\
        & \multirow{2}{*}{Siamese} & \cmark & 0.5523 & 0.6399 \\
        & & \xmark & 0.5313 & 0.5264 \\
        \midrule
        \multirow{3}{*}{\cmark} & TriDeNT & \cmark & \B 0.9159 & 0.9080 \\
        & \multirow{2}{*}{Siamese} & \cmark & 0.8931 & 0.8274 \\
        & & \xmark & 0.8914 & 0.8506 \\
        \bottomrule
    \end{tabular}
    }
    \caption{}
    \label{tab:kmeans-pannuke}
    \end{subfigure}
    \begin{subfigure}{0.42\textwidth}
        \includegraphics[width=\textwidth, trim={ 0 0.4cm 0 0 }, clip]{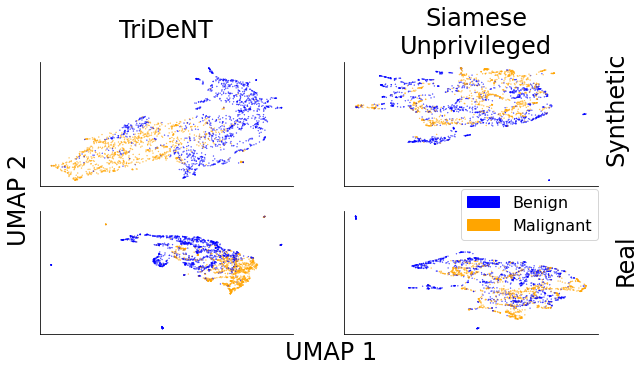}
        \caption{}
        \label{fig:umap-figure}
    \end{subfigure}
    \caption{(a) Evaluation of representations from models trained on the real PanNuke dataset and the synthetically generated nuclear segmentations paired with the NCT dataset. (b) UMAP projections, coloured by PanNuke label.}
    \label{fig:enter-label}
\end{figure}

\subsection{Training With Synthetically Generated Privileged Data Outperforms Training on the Real Source Dataset}

Another key use case for synthetically generated privileged information is where only a small amount of data is available. In Figure \ref{tab:kmeans-pannuke}, we evaluate models trained on the small PanNuke dataset by performing unsupervised $k$-means clustering with 2 clusters to determine binary labels. Trained with synthetically generated privileged information, TriDeNT was able to accurately make completely unsupervised classifications. For all other methods, performance is little better than random, implying that neither the PanNuke train set nor unprivileged training on the larger dataset is sufficient for good performance. This can be intuitively seen from Figure \ref{fig:umap-figure}, where TriDeNT differentiates between the two classes more than unprivileged methods, with qualitative similarities between the projections of real and synthetic models. This shows that this approach could be used to better analyse small datasets from rare diseases or underrepresented populations, which may not have sufficient labels for classifier training, either for basic research or in a clinical setting. Supervised evaluation produces similar results, with the TriDeNT model trained on NCT with synthetically generated PanNuke patches outperforming training directly on the real PanNuke dataset.

It is notable that models performed well which were trained with synthetically generated privileged information from small datasets such as PanNuke and SHIFT, which have generator train sets of 2205 and 665 images respectively. Figure \ref{fig:realvsynthetic} demonstrates that the reduction in error from training on the larger synthetic dataset compared to the real smaller dataset is more than 400\% and 200\% for PanNuke and SHIFT respectively.

\section{Discussion and Conclusions}

\subsection{Synthetically Generated Privileged Data Improves Model Performance in Both In- and Out-of-Distribution Settings}

We have established that the performance of single-input models is significantly improved by training with synthetically generated paired data. This can be considered as a form of implicit multi-objective learning, as the image generation model has been optimised to perform a different task to the self-supervised pretext task, and this provides the commensurate benefits. Despite this, the required computational power at any point is no different to standard Siamese/TriDeNT learning, as the images are generated prior to the self-supervised training. In many cases, these generative models are already available and could be used to create images without being trained. This means that these methods can be accessible to clinicians and researchers without large-scale resources, and enable the application of self-supervised privileged training to datasets which are otherwise intractable due to a lack of paired data.

\subsection{Clinical Impact and Future Work}

This work was primarily motivated by two use cases: unpaired datasets, and small/rare datasets. Many state-of-the-art datasets do not have accompanying routine data, and this approach allows insights from those data to be integrated into routine data. Furthermore, many datasets are very limited in size or scope, particularly for rare diseases or those affecting underrepresented populations. This approach could enable retrospective reanalysis of routine data from rare samples using insights from current state-of-the-art data, meaning routine data from both rare and common cases can be fully utilised. This is especially prescient, given that TriDeNT was shown to significantly outperform comparable methods for few-shot applications \cite{farndale2023trident}.

We have limited the scope of this study to the case of a single paired input, but further improvements could be made with the addition of multiple paired datasets. This could be easily achieved by applying several generative models to the same large primary dataset. We have also not considered training with the real data shuffled into the synthetically generated dataset, which could further improve performance.

\section*{Acknowledgements}

LF is supported by the MRC grant MR/W006804/1, CW and RI acknowledge support from CRUK A24450. RI is supported by EPSRC EP/S0300875/1 and Wellcome 221786/Z/20/Z. KY acknowledges support from EP/R018634/1, BB/V016067/1. We acknowledge support from A. Ammar, J. Hay, H. Morgan, C. Nixon, J. Quinn, J. Edwards and the Glasgow Tissue Research Facility.

\bibliographystyle{unsrt}
\bibliography{refs}

\setcounter{figure}{0}
\renewcommand{\figurename}{Fig.}
\renewcommand{\thefigure}{S\arabic{figure}}
\setcounter{table}{0}
\renewcommand{\tablename}{Table}
\renewcommand{\thetable}{S\arabic{table}}
\setcounter{section}{0}
\renewcommand{\thesection}{S\arabic{section}}

\clearpage

\section*{Supplementary Material}

\begin{table}[h]
    \centering
    \caption{Datasets}
    \label{tab:datasets}
    \scalebox{0.65}{\begin{tabular}{ccccccc}
    \toprule
         Name & \makecell{Tissue\\Type} & \makecell{Train/Test\\Split} & Training & Evaluation & Purpose & Task \\
    \midrule
         NCT \cite{kather_dataset} & Colorectal & 100,000/7,177 & \cmark & \cmark & \makecell{Cancer Tissue Analysis} & \begin{tabular}{@{}c@{}}\\\\\end{tabular}Tissue Type Classification \\
         \midrule
         \multicolumn{7}{l}{\footnotesize\textbf{Note:} Nine tissue types: adipose (ADI), background (BACK), debris (DEB), lymphocytes (LYM), mucus (MUC), smooth} \\
         \multicolumn{7}{l}{\footnotesize muscle (MUS), normal colon mucosa (NORM), cancer-associated stroma (STR), colorectal adenocarcinoma epithelium (TUM)} \\
    \midrule
        Camelyon17 \cite{bandi2018detection} & Lymph Node & 179,394/146,722 & \xmark & \cmark & \makecell{Out-of Distribution\\Metastasis Detection} & \makecell{Out-of-Distribution\\Binary Classification} \\
        \midrule
        \multicolumn{7}{l}{\footnotesize\textbf{Note:} Train set from 3 centers, validation set from center 4, and test set from center 5} \\
    \midrule
         PanNuke \cite{gamper2019pannuke} & Pan-Cancer & 2,205/4,373 & \cmark & \cmark & Nuclear Segmentation & \begin{tabular}{@{}c@{}}\\\\\end{tabular}Neoplastic Cell Detection \\
         \midrule
         \multicolumn{7}{l}{\footnotesize\textbf{Note:} As in \cite{huang2023visual,farndale2023trident}, patches are labelled as benign if no neoplastic cells are present, and malignant if the patch contains at least} \\
         \multicolumn{7}{l}{\footnotesize 10 cells and 30\% of the cells are neoplastic.} \\
    \midrule
         Synthetic IHC \cite{walsh2022ensuring} & Breast & 93,952/- & \cmark & \xmark & \makecell{Virtual Restaining\\Brightfield AE1/AE3} & - \\
         \midrule
         \multicolumn{7}{l}{\footnotesize\textbf{Note:} Originally 23,488 $448\times448$px patches, which were split into 4 smaller patches} \\
    \midrule
         SHIFT \cite{burlingame2020shift} & Prostate & 665/11,593 & \cmark & \xmark & \makecell{Virtual Restaining\\IF pan-CK} & - \\
         \midrule
         \multicolumn{7}{l}{\footnotesize\textbf{Note:} All 12,258 patches used in SSL training. Generator trained on only 665. Only 4 patients' data in dataset.} \\
    \bottomrule
    \end{tabular}
    }
\end{table}

\begin{table}[h]
\centering
\caption{Results for models trained with synthetically generated nuclear segmentations as privileged information for the NCT dataset. Unlike in the results in the main text, these segmentations simply mask the background, leaving the nuclei. There are no type annotations. Evaluation is performed on the NCT and Camelyon datasets.}
\label{tab:3_branch_image_nuclei}
    \scalebox{1}{\begin{tabular}{ccccc}
    \toprule
        Loss & Method & Privileged & NCT & Camelyon \\
        \midrule
          \multirow{3}{*}{VICReg}
          & TriDeNT & \greycell\cmark & \greycell\B 0.9280 & \greycell\B 0.8330 \\
          & \multirow{2}{*}{Siamese} & \greycell\cmark & \greycell 0.7778 & \greycell 0.7871 \\ 
          & & \xmark & 0.8937 & 0.7056 \\
        \midrule
         \multirow{3}{*}{InfoNCE} & TriDeNT & \greycell\cmark & \B\greycell 0.9334 & \greycell 0.6859 \\
         & \multirow{2}{*}{Siamese} & \greycell\cmark & \greycell 0.9058 & \greycell\B 0.8878 \\ 
         & & \xmark & 0.9047 & 0.6205 \\
        \bottomrule
    \end{tabular}}
\end{table}

\begin{figure}
    \centering
    \includegraphics[width=0.6\textwidth]{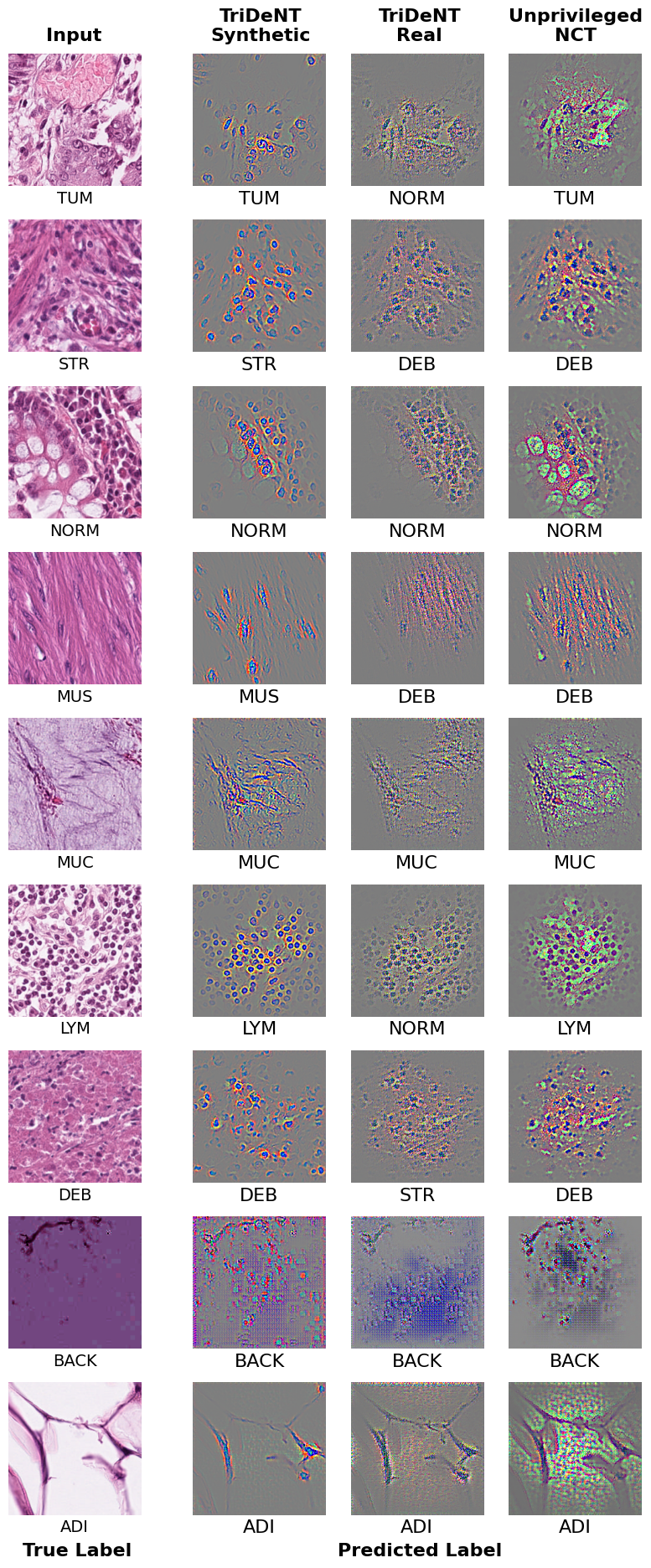}
    \caption{Representative samples from each NCT tissue class with Guided-GradCAM activation maps. True and predicted labels are shown for each classifier.}
    \label{fig:full-guided-gradcam}
\end{figure}

\end{document}